\documentclass[runningheads]{llncs}
\usepackage{graphicx}
\usepackage{amsmath}
\usepackage{amsfonts} 
\usepackage{caption}
\usepackage{tabularx}
\usepackage{float}

\usepackage{subcaption}
\DeclareMathOperator*{\argmin}{arg\;min} 

\begin{document}
\title{Depthwise Discrete Representation Learning}
%
%
\author{Iordanis Fostiropoulos}
%
%
\institute{University of Southern California \\
\email{fostirop@usc.edu}}
\maketitle              
\begin{abstract}

 Recent advancements in learning Discrete Representations as opposed to continuous ones have led to state of art results in tasks that involve Language, Audio and Vision. Some latent factors such as words,phonemes and shapes are better represented by discrete latent variables as opposed to continuous. Vector Quantized Variational Autoencoders (VQVAE) have produced remarkable results in multiple domains. VQVAE learns a prior distribution $z_e$ along with its mapping to a discrete number of $K$ vectors (Vector Quantization). We propose applying VQ along the feature axis. We hypothesize that by doing so, we are learning a mapping between the codebook vectors and the marginal distribution of the prior feature space. Our approach leads to 33\% improvement as compared to prevous discrete models and has similar performance to state of the art auto-regressive models (e.g. PixelSNAIL). 
 We evaluate our approach on a static prior using an artificial toy dataset (blobs). We further evaluate our approach on benchmarks for CIFAR-10 and ImageNet.

\keywords{Vector Quantized Variational Autoencoder \and  Generative Models}
\end{abstract}
\section{Introduction}

Learning Representations according to Bengio et al. \cite{bengio2013representation}, involves learning the explanatory factors of the observed input. In this paper we refer to those factors as latent factors. Learning good representations leads to superior performance in both supervised and unsupervised tasks across several domains. 



\subsection{Discrete Representations}
Variational Auto-Encoders(VAE) have seen a wide use in a multiple domains \cite{kusner2017grammar,xu2017variational}. VAEs have good training performance in an unsupervised setting as opposed to other unsupervised learning techniques such as GAN \cite{goodfellow2014generative} that suffer from issues like mode collapse. There are multiple interpretations of a VAE. Kingman et al \cite{kingma2013auto} originally defined them as probabilistic generative models. 
Subsequent work expanded on the original probabilistic model interpretation and favored the interpretation of VAEs as information bottlenecks \cite{higgins2017beta}. InfoVAE\cite{zhao2019infovae} further expanded on the above idea and aim at maximizing the information between latents and the input. 
Several previous works have introduced improvements on VAEs for both the continuous and discrete case \cite{jang2016categorical}. 


Learning discrete representations can lead to improved performance in several tasks for which the underlying latent factors are discrete \cite{kusner2017grammar}. Consider as an example a derivative work of WaveNet\cite{chorowski2019unsupervised}, in which the discrete embedding are paired with a powerful auto-regressive model.
The learned discrete representations were able to capture features such as phonemes of speech in an unsupervised manner. 

 A discrete embedding can provide a compressed latent space with similar reconstructions to its continuous counterparts \cite{chen2018learning}. Current work has shown exceptional results of discrete models in generative tasks when they are paired with a powerful auto-regressive model. This is because the discrete representations are more efficient to pair with an auto regressive model as compared to their continuous counterparts \cite{razavi2019generating}\cite{dieleman2018challenge}

Interpreting discrete representations is much more intuitive for humans than continuous values. We have a discrete view of the world. We interpret symbolic reasoning and language as discrete variables. Thus it is much easier to understand and interpret the learned representations in the discrete case when compared to the continuous. 



One downside in learning discrete representations is that gradient calculation and accurate attribution is only possible via approximation. There are many approaches in approximating the gradients. The most common one being straight-through which is calculating the gradients with respect to the input in the discrete layer \cite{bengio2013estimating} 
There are several works that can learn discrete latent variables \cite{makhzani2015adversarial,fedus2018maskgan}. In this work we focus on VQVAE \cite{van2017neural}



Vector Quantizers (VQ) \cite{gray1984vector} have been widely used for audio compression. They can be interpreted as information bottlenecks. The objective of VQ is to learning a mapping between a set of vectors in a space, such that the vectors are the mean value of the data closest to them. 
VQVAE uses this objective to jointly train a VQ as a density estimator of the prior as well as train a prior such that it can be best reconstructed after being quantized. 
Vector Quantization is used in domains outside Auto Encoders and there is still active research in the area \cite{villmann2017can}. There have been learning paradigms around them \cite{kohonen1982self}\cite{sato1996generalized} and recent research has purposed improvements for training them at scale \cite{martinez2018lsq++}.




\subsection{Problems with Vector Quantization}

As noted in previous work, VQ is difficult to optimize for high dimensional feature spaces \cite{lee1995fast}. For the training objective that we use in this paper, finding an optimal mapping was sensitive to the dimension of the feature space as well as the number of underlying latent factors.

A natural approach would be to increase the number of vectors we use to perform density estimation to the same number of underlying latent factors. When the number of underlying latent factors is far more than what is computationally feasible by a VQ this approach leads to no advantage in compression and only to further computational performance degradation. 

Thus, we identify that the problem with using VQ as density estimators is two-fold. It is challenging to optimizing VQ for high dimensional spaces. Increasing the number of vectors does not lead to a reconstruction advantage when the underlying latent factors are far more than can be effectively represented by a codebook.

For some applications of VQ the problems outlined might not be present. Consider we are using VQ directly to an image. The feature axis of an image would be the number of pixel channels, thus it is 3 (for "RGB"). Similarly for audio that signal amplitude is the only feature. Moreover the probability density function of the input signal would be uniformly distributed. Thus VQ is expected to perform reasonably well.






\subsection{Contribution}
We make the following contributions with this paper:
\begin{itemize}
    \item When VQ is interpreted as a density estimator, we identify and mitigate issues on its compression performance. Those issues are caused by the data dimensionality as well as the number of latent factors. 
    \item We introduce Depthwise Vector Quantization; multiple independent vector quantizers are applied along the feature axis. 
    \item We improve current state of art benchmarks in discrete representation models that are of similar performance to auto-regressive state of art counterparts.
\end{itemize}

In detail, we perform an analysis of VQ on high dimensional feature spaces with a fixed prior and with a known number of latent factors. We trained a vanilla VQ in an unsupervised learning setting. 
We conclude that when the number of latent factors is significantly large, a larger codebook will provide no reconstruction advantage for the training objective we are set at optimizing. 
Moreover we find that for feature spaces that have larger than 16 dimensions (as also noted by previous works) VQ fails to converge at an optimal solution, regardless of the codebook size. For high dimensional feature spaces the compression loss converges to the same bound, regardless of the growth on the code-book size.


We introduce a depthwise vector quantization technique. We improve performance by splitting the data into N disjoint sets in feature space. We see substantial improvements in compression rates for all of our experiments that consider a static prior.

Lastly we apply depthwise vector quantization when the prior is learned and not static. We are able to obtain better reconstructions than previous work. We hypothesize part of the reason is that the codebooks learn marginal distributions and encourage independence between the features. 
Depthwise Vector Quantization is able to improve on compression benchmarks from previous approaches. We obtain  33\% improvement in reconstructions as compared to VQVAE on CIFAR-10.


\section{Background}

In this paper, we build directly on top of two previous works. Vector Quantizers and Vector Quantized Auto-Encoders.

\subsection{Vector Quantization}

Given an input vector $x\in \mathbb{R}^{D}$, where $D$ is the dimension of the input feature space. A Vector Quantizer learns a mapping between $x\in \mathbb{R}^D$ and a set of codebook vectors $e \in {R} ^{K\times D}$, where $K \in \mathbb{Z}_+ $ is the number of unique vectors. We refer to the set of codebook vectors as $C=\{e_1\dots e_K\}$, a \textbf{codebook}.  

The quantization of an input $x_i$ is calculated with respect to some distance function $d$

$$
z_i = \argmin_{j\in K}\;d(x_i,e_j)
$$

Where $z_i \in [1,K] \subset \mathbb{Z}_+ $ is the codebook index of the centroid vector closest to $x_i$. The objective function minimizes the distance of each input vector $x_i$ to the closest vector. It updates $e$ to minimize the distance with the closest $x_i$. The optimization step can be summarized as:

\begin{equation}
\min_e\; \sum_{i\in |x|} d(x_i,e_{z_i}) 
\label{eqn:vq_obj}
\end{equation}

For our experiments and the remainder of this paper we consider the Euclidean Distance between expectations (original input $x_i$) and the quantized vectors. 

The \textbf{optimal mapping} is when the quantized vectors are the average of all the data in its quantization region. As $D$ grows it becomes intractable to find the optimal mapping for vector quantization. As long as the distance function $d$ is differentiable we can minimize the objective \eqref{eqn:vq_obj} via gradient-based optimization techniques. VQ can also be thought as K-means clustering in this setting.

VQ is sensitive to the dimension $D$ of the input as well as the number $K$ of quantization vectors $e$. A larger number of vectors could lead to improved reconstructions, although this is not always the case for our training objective. Hence it is not reasonable to choose a really large K. 


\subsection{Vector Quantized Variational Auto-Encoder}

Variational Auto-Encoders are composed of the parametric encoder $q_\phi(z|x)$, a prior $p(z)$ and a decoder network $p_\theta(x|z)$. In VQVAE the latent space $z$ is discrete. Given a continuous vector $z_e$, VQ returns a quantized tensor space $z_q$. $z_q$ is the concatenation outcome of all quantized vectors from $z_e$. The prior $z_e$ is learned rather than being static. The Vector Quantizer is trained by decreasing the distance between the quantization vectors $e\in R^D$ and the learned prior. Both the VQ and the prior have the same objective of reducing the distance between them. Thus the prior can be interpreted as trying to generate densities that can be most efficiently quantized and the vector quantizer is estimating those densities as efficiently as possible. 

This is reflected by the objective function:

\begin{equation}
    L=\underbrace{\log{p_\phi(x|z_q)}}_{\text{Reconstruction Loss (i)}} + \underbrace{\beta\;d(sg(z_e),z_q)}_{\text{Commitment Loss (ii)} } + \underbrace{d(z_e,sg(z_q))}_{\text{VQ Loss (iii)}}
    \label{vqvae_obj}
\end{equation}

Reconstruction Loss optimizes both the encoder and decoder. This is done using a straight-through gradient estimator \cite{bengio2013estimating} between the discrete outcomes $z_q$ and $z_e$. During back-propagation the gradients at $z_q$ are applied to $z_e$ and the original gradients at $z_q$ are ignored. Because the gradients at $z_q$ are ignored, straight-through estimator doesn't attribute any of the reconstruction loss between encoder and decoder to the vector quantization layer. 
Since both $z_q, z_e \in \mathbb{R}^D$ have the same dimension $D$, the gradients are good estimator in training the encoder network.

Commitment Loss optimizes the learned prior with $\beta$ being the commitment coefficient. This term regularizes large updates on the learned prior. This would be because on each forward pass the encoder output could be discretized differently, making it challenging to learn a good mapping between $z_q$ and $z_e$

VQ Loss trains a codebook $C$ to optimize the mapping between $z_q$ and $z_e$ (same as equation \ref{eqn:vq_obj}). For both (ii) and (iii) we use the $sg$ stop gradient function. During back-propagation returns 0 and during a forward pass is the identity function. Thus we can optimize independently the VQ via term (iii) and the prior via term (ii).

\begin{figure}
\includegraphics[width=\textwidth]{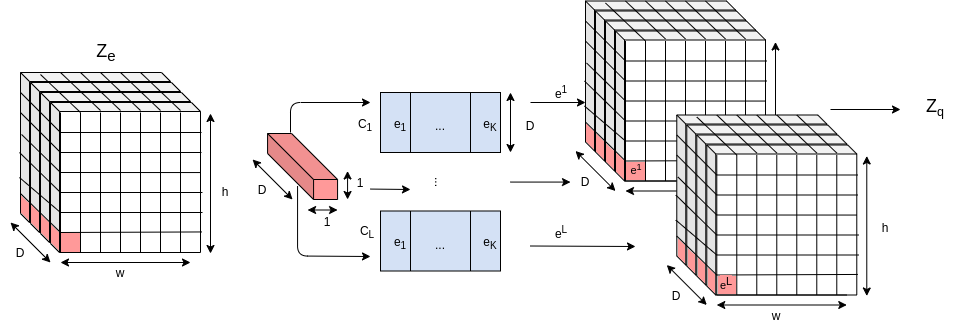}
\caption{VQVAE quantization process \cite{van2017neural}} \label{fig:vqvae}
\end{figure}

In the original implementation of VQVAE $L$ number of codebooks can be used. Although this is not always the case in practice as we outline in Section \ref{sec:feature_indepedence}. The overall architecture of the original implementation of VQVAE is shown in Figure \ref{fig:vqvae}. 

In this example we consider $w\times h$ spatially correlated input vectors with a feature space $D$. $z_e \in \mathbb{R}^{w\times h \times D}$ can be the result of a non-linear encoder network such as a Convolutional Neural Network (CNN), the above can generalize to any type of encoder network. All the codebooks are trained on the same prior $z_e$, the outcome of the quantization from each codebook can be (but not limited to) concatenation.

\section{Depthwise VQ}

We propose depthwise vector quantization. We define a latent space $z_e \in \mathbb{R}^{D \times w \times h}$ resulting from a learned or static prior. A disjoint set $\hat{z_e}$ of $L$ \textbf{feature vectors}  contains slices of size $D_i$ of the prior vector $z_{e}$ along the feature dimension such that
\begin{equation}
    \hat{z_{e}}=\{{z_{e}}_{[1:D_1]},\dots,{z_{e}}_{[D_{L-1}:D_L]}\}\;\;\text{with}\;\;|\hat{z_{e}}|=L\;\;
\end{equation}
We learn a mapping for each codebook $C_i$ in our a set of \textbf{codebooks} $C=\{C_1,\dots,C_L\} \in \mathbb{R}^{L\times K\times D}$ with it's pairwise slice from $\hat{z_{e}}$. For simplicity we will denote $z_{[0:D_n]}=z_n$ in the remainder of this section.

Consider the example of codebooks ${C_1,C_2,C_3}$ and slices $z_1,z_2,z_3$, our objective is to learn a mapping $z_n \mapsto C_n=z_{q_n}$, thus each $z_n$ results in a quantized counterpart $z_{q_n}$. This results in $z_1=z_{q_1} \dots$, and $[z_{q_1},z_{q_2},z_{q_3}]=z_q$. The resulting tensor $z_q$ can be (but not limited) the result of concatenation such that $z_{q}=\cup_i^L{z_{q_n}}$

The objective from Equation \ref{vqvae_obj}, for a learned prior is then changed to:

\begin{equation}
    L=\underbrace{\log{p_\phi(x|z_q)}}_{\text{Reconstruction Loss (i)}} + \underbrace{\beta\;\sum_i^{L}{d(sg(z_i),z_{q_i})}}_{\text{Commitment Loss (ii)} } + \underbrace{\sum_i^L{d(z_{i},sg(z_{q_i}))}}_{\text{VQ Loss (iii)}}
    \label{dvqvae_obj}
\end{equation}

\begin{figure}
\includegraphics[width=\textwidth]{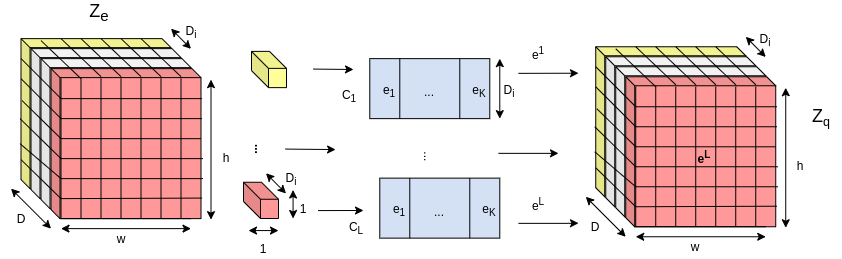}
\caption{Depthwise Vector Quantization, the data is split along the feature axis.} \label{dvq}
\end{figure}

The overview of our proposed model is shown at Figure \ref{dvq}, where given a prior we split the tensor into slices of size $D_i$ along the feature axis. We pass the disjoint slices to their pairwise codebooks. This results into $w\times h \times L$ discrete latent space $z$.

\subsection{Feature Independence}\label{sec:feature_indepedence}

The codebooks work as density estimators \cite{gray1997vector} on marginal distributions as opposed to the joint feature space.
We hypothesize that by encouraging the independence between the features we allow for an exponentially higher number of latent factors to be expressed with a linear growth on the number of codebooks. DVQ can theoretically express $K^L$ factors, where $K$ is the number of codebook vectors (discrete possible outcomes) and $L$ is the number of codebooks.

In contrast, when multiple codebooks are trained on a joint distribution it would approximate a linear growth of factors with the linear growth of codebooks. Consider the example of a VQ applied on $z_e$ jointly (as in Figure \ref{fig:vqvae}). 

In the ideal case that all $L$ codebooks learn a perfect mapping for $z_e$, the codebook vectors would be identical. Even if the discrete indexes are different such that the vectors $e_i=e_j$ with $i\neq j$, the quantized latent space would be the result of concatenating $[e_i,e_j]=[e_i,e_i]$. There would be no additional benefit in this case in using more than one codebook. The total number of possible discrete outcomes would be $K$, regardless of the number $L$ of codebooks.

In reality learning the ideal mapping between the multiple codebooks and the input $z_e$ is intractable for the problem space we are working. The codebooks learn different approximation of the same $z_e$. Each codebook will converge to a slightly different mapping. Thus the expressiveness of this approach would be much better than $K$ but still far worse than $K^L$.






Applying VQ depthwise (along the feature axis) provides the benefits of learning a better density approximation and reducing the intractability of VQ for large $D$ dimensions. 

The first argument is a consequence of the density approximation learned independently. It is allowing us to express an exponential number of latent factors given $L$ codebooks with $K$ vectors we can express $K^L$ factors. With a linear increase in the number of parameters (in this case codebook vectors) we have to optimize, we obtain exponential increase in the number of latent factors that can be expressed. 

The second argument is a consequence of dividing $D$ into $L$ number of slices. Multiple codebooks do not degrade performance at the same rate as a single larger codebook does. Multiple codebooks can allow for better parallelism, since the quantization can be performed independently on different feature subsets. 

Consider the example of having infinitely more latent factors than we can effectively learn and a single code-book with a large $K$. Such codebook will be able to express only $K$ latent factors. The minimum vector look up will take $O(K)$ for each training step. Theoretically we can express the same number of latent factors with $K_{DVQ}=\sqrt[L]{K}$, the performance benefits in expressiveness become apparent in using multiple feature independent codebooks as opposed to a single large codebook. This is also validated by our experiments when learning a static prior.

\section{Experiments}

Throughout the experiments we keep the implementation of VQ the same as the vanilla VQVAE. We do not consider additional changes as introduced by derivative works\cite{roy2018theory}. Our work does not coincide with the derivative works that propose changes on different aspects of VQ as compared to our work. Our work can be applied irrespective of those changes.

We divide our experimental section into two categories of experiments. Ones done on a static prior and ones done on a learned prior. For the static prior experiments we used artificially generated blob data-sets. The rationale was that we could control the number of latent factors (clusters of blobs). When the prior was learned we considered two datasets CIFAR-10 and ImageNet.

\subsection{Static Prior}

We generated the static prior using blob dataset \cite{scikit-learn} which we randomly initialized as Gaussian mixtures. We varied the number of Gaussian (latent factors) $N_{\text{G}}$ with $\mu\in [-1,1]^D$ and $\sigma=0.01$. We used a uniform random initializer for the codebook centroids with $\mu = 0$ and $\sigma=1$. We used the artificial dataset to train 3 models; Depthwise VQ (\textbf{DVQ}), single codebook VQ (\textbf{VQ}), single codebook VQ with a large codebook size $K < N_{\text{G}}$ (\textbf{$\text{VQ}^{K+}$}). 

For \textbf{DVQ}, we split the feature axis on the dataset in $L$ sets that were of equal size $D_i$ and disjoint. We then trained $L$ independent codebooks on each subset of features. Note that $L\times D_i=D$. For \textbf{VQ} we trained a single codebook $L=1$ on the entire feature space $D$. We kept the same number of codebook vectors $K$ between \textbf{DVQ} and \textbf{VQ}. For \textbf{VQ$^\text{K+}$} we trained a single codebook significantly larger than the previous experiments but still smaller than the true number of underlying factors. 

$$
K_{DVQ}=K_{VQ}<<K_{VQ^{K+}}<N_G
$$

Our goal was to evaluate against a baseline and investigate if the application of DVQ along the feature axis using multiple codebook results in a better performance as opposed to using a single large codebook (that can represent significantly more factors with higher accuracy). For this reason, we varied $K$, $N_G$ and $D$ and denote their values for each individual experiment.

We cross-validated our results by performing Monte-Carlo sampling. We trained for the same number of training steps that we ensured would lead for the loss to converge for all models. In all of the reported results and figures we take the mean of the Monte-Carlo sampling, with $N=10$ repetitions.

\begin{figure}
\begin{subfigure}{.5\textwidth}
\includegraphics[width=\textwidth]{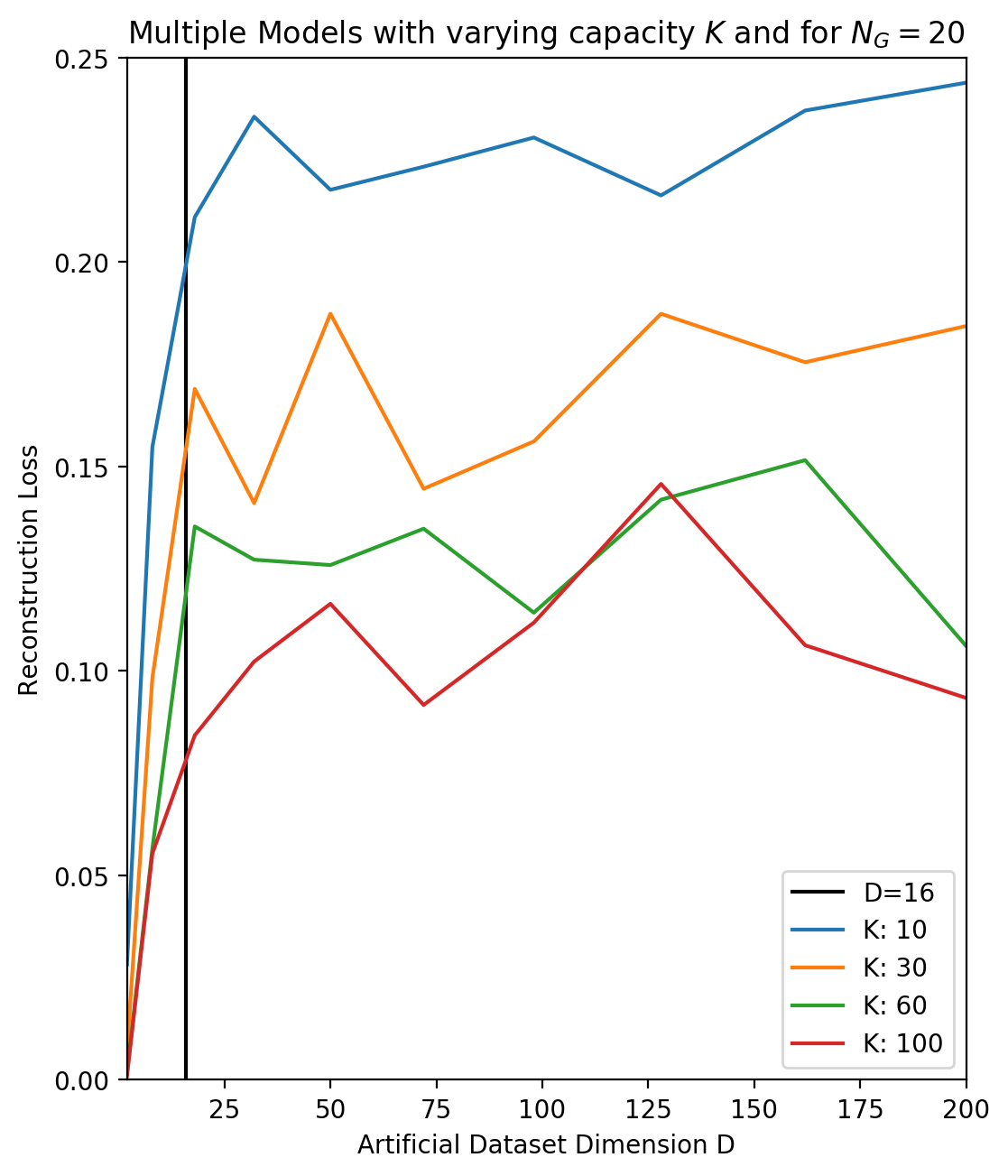}
\caption{We observe the effect on the loss as we increase the feature space dimension} \label{fig:kvsd}
\end{subfigure}
\begin{subfigure}{.5\textwidth}
\includegraphics[width=\textwidth]{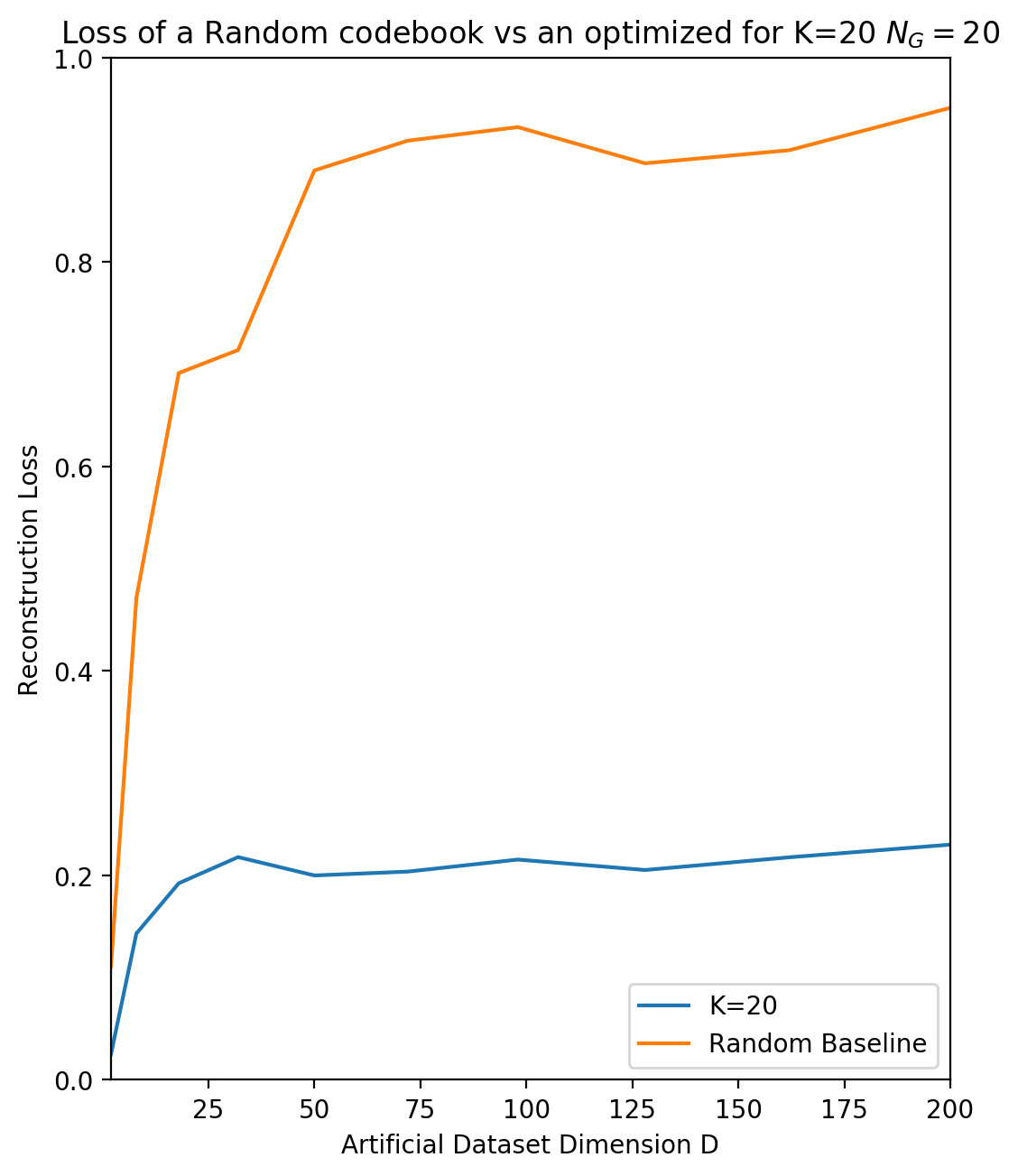}
\caption{We compare a single codebook with a random baseline for different dimensions of input } \label{fig:kvsrandom}
\end{subfigure}
\caption{Results of VQ on a static prior for different size codebooks.}
\end{figure}

High dimensional latent space results in difficulty in training jointly. We found in agreement with previous work \cite{raginsky2006estimation} that for $D<16$, the problem of finding ideal mapping is tractable even as an optimization problem and using a larger number of density estimation factors ($K$) can lead to improvement in loss. For D larger than 16 regardless of the code-book size all codebooks converge to the same compression rate (reconstruction loss). Figure \ref{fig:kvsd}, shows the reconstruction loss for varying capacity quantizers as we increase the feature space dimension. It is worth noting 2 things. Firstly when the number of centroids is far smaller than the true number of latent factors, the performance greatly degrades (however doesn't become worse than random Figure \ref{fig:kvsrandom}). When $D<16$ the larger number of centroids provides better convergence guarantees. For larger dimensions the density estimation becomes intractable and the centroids only provide an approximation of the true latent factors. 
We conducted experiments for $K=[10,20,\dots,100]$ and for different dimensions. We report the Mean Loss on the test dataset for experiments performed on different dimensionality feature spaces. There are approaches to mitigate to some extend the above issue for large dimensions\cite{shen2006adaptive}. Most approaches however are approximation based.

\begin{figure}
\begin{subfigure}{.33\textwidth}
\includegraphics[width=\textwidth]{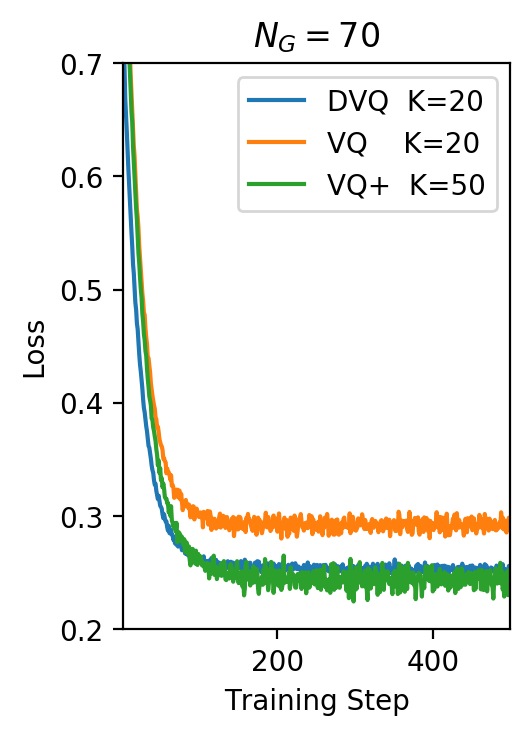}
\end{subfigure}
\begin{subfigure}{.32\textwidth}
\includegraphics[width=\textwidth]{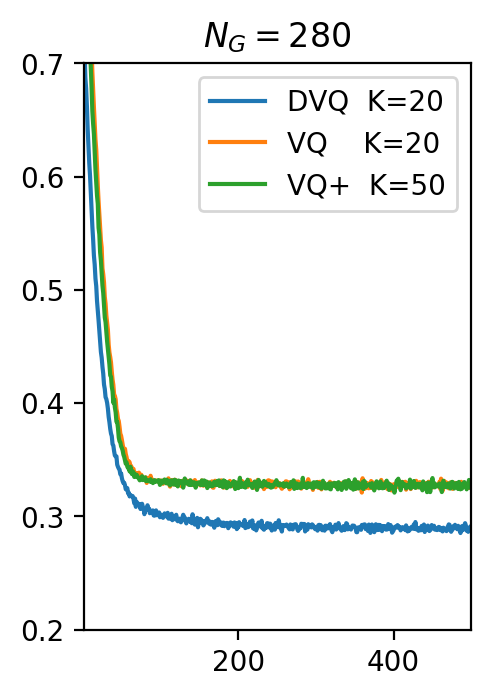}
\end{subfigure}
\begin{subfigure}{.33\textwidth}
\includegraphics[width=\textwidth]{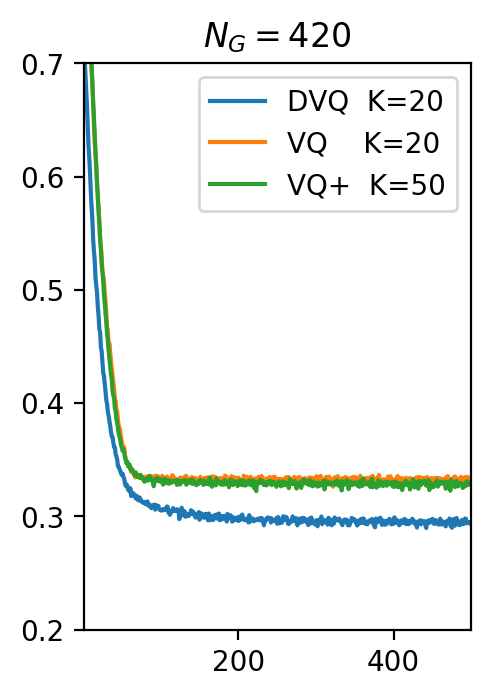}
\end{subfigure}
\caption{Comparison between DVQ and VQ for different number of latent factors $N_G$ Increase in the codebook size doesn't improve loss when the number of underlying factors are beyond the codebook capacity. $DVQ$ performs better than same capacity single codebooks and at par or better than larger capacity codebooks.} 
\label{DVQ_VQ}
\end{figure}

\subsubsection{Latent Factors}

When we compare multiple depthwise codebooks as opposed to a single codebook (with the same number of centroids), we obtain an improved reconstruction loss. We chose $K$ to be smaller than the number of underlying latent factors. For the experiments presented in Figure \ref{DVQ_VQ} we chose $K=20$ with the true number of factors to be $N_G=70$. A Single Larger codebook of $K=50$ still under-performs as compared to $K=20$. The disparity between DVQ and VQ becomes even more apparent as the number of factors becomes prohibitively large (as it could be the case for a natural dataset e.g. CIFAR-10). From Figure \ref{DVQ_VQ}, where for a large number of underlying latent factors (e.g. $N_G=420$), $K$ becomes irrelevant. However training independently on the feature space multiple codebooks still provides a performance advantage. Again there is a limit to the performance boost we can get since theoretically is not possible to estimate the density of significantly larger number of factors than codebook vectors.

\subsection{Learned Prior}

\begin{figure}
\includegraphics[width=\textwidth]{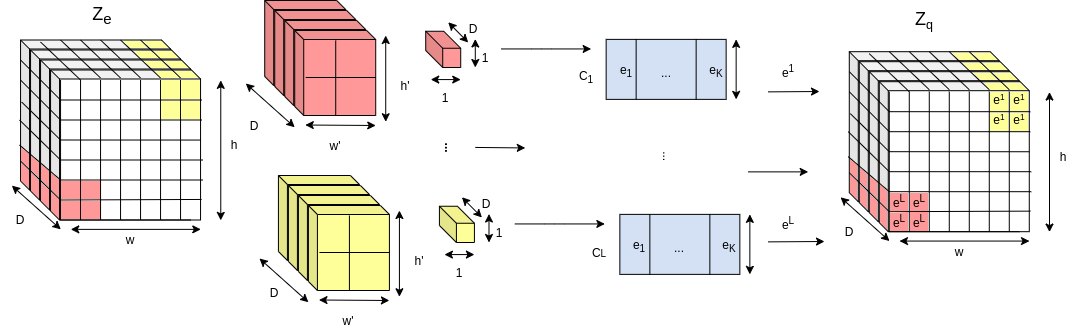}
\caption{Splitting and quantization process for spatially independent codebooks} \label{fig:svq}
\end{figure}

We evaluate our approach on CIFAR-10 and ImageNet. We also further evaluate our hypothesis by considering VQ trained on spatially independent codebooks (\textbf{SVQ}). We use the same encoder and decoder architectures when comparing across different models on a given dataset. The architecture of our encoder/decoder is identical to VQVAE \cite{van2017neural}.  \textbf{SVQ}, is used as a baseline to demonstrate that the advantage of our approach is in our hypothesis. We train spatially independent codebooks. We do this by splitting along the spatial dimension (e.g. width and height) of $z_e$. This is a bad assumption for an image, since pixels are spatially dependent. The full approach and how it compares to \textbf{DVQ} and \textbf{VQ} can be seen at Figure \ref{fig:svq}. For all of our implementations we use the same $D=250$, $K=512$, $L=8$ $\text{batch}=128$ and we train for approximately 100000 steps for all models. 

For the DVQ case we observed that performs best when we consider multiple features (filters) at once $|D_i|>2$. We obtain improved reconstructions using our approach. DVQ surpasses in performance SVQ, agreeing with our hypothesis that it is the independent feature learning that improves the reconstructions. Moreover spatial features in an image are highly dependent so our assumption on independence doesn't hold true (and hence no additional performance benefit). Moreover there is performance degradation by the naive assumption.

We conclude that the benefit of DVQ is not only because we are training independent codebooks, but because those codebooks encourage the encoder to generate independent densities in the feature axis. Moreover our assumption on the feature independence is useful in increasing the performance.

DVQ surpasses in all metrics it's counterparts. DVQ does as well as Flow Based Generative Models and Autoregressive models. PixelCNN 3.14 bits/dim  \cite{oord2016pixel} Flow++ 3.08 bits/dim \cite{ho2019flow++} as compared to 3.15 bits/dim for our approach on CIFAR-10. 

\begin{figure}
\begin{subfigure}{.3\textwidth}
\includegraphics[width=\textwidth]{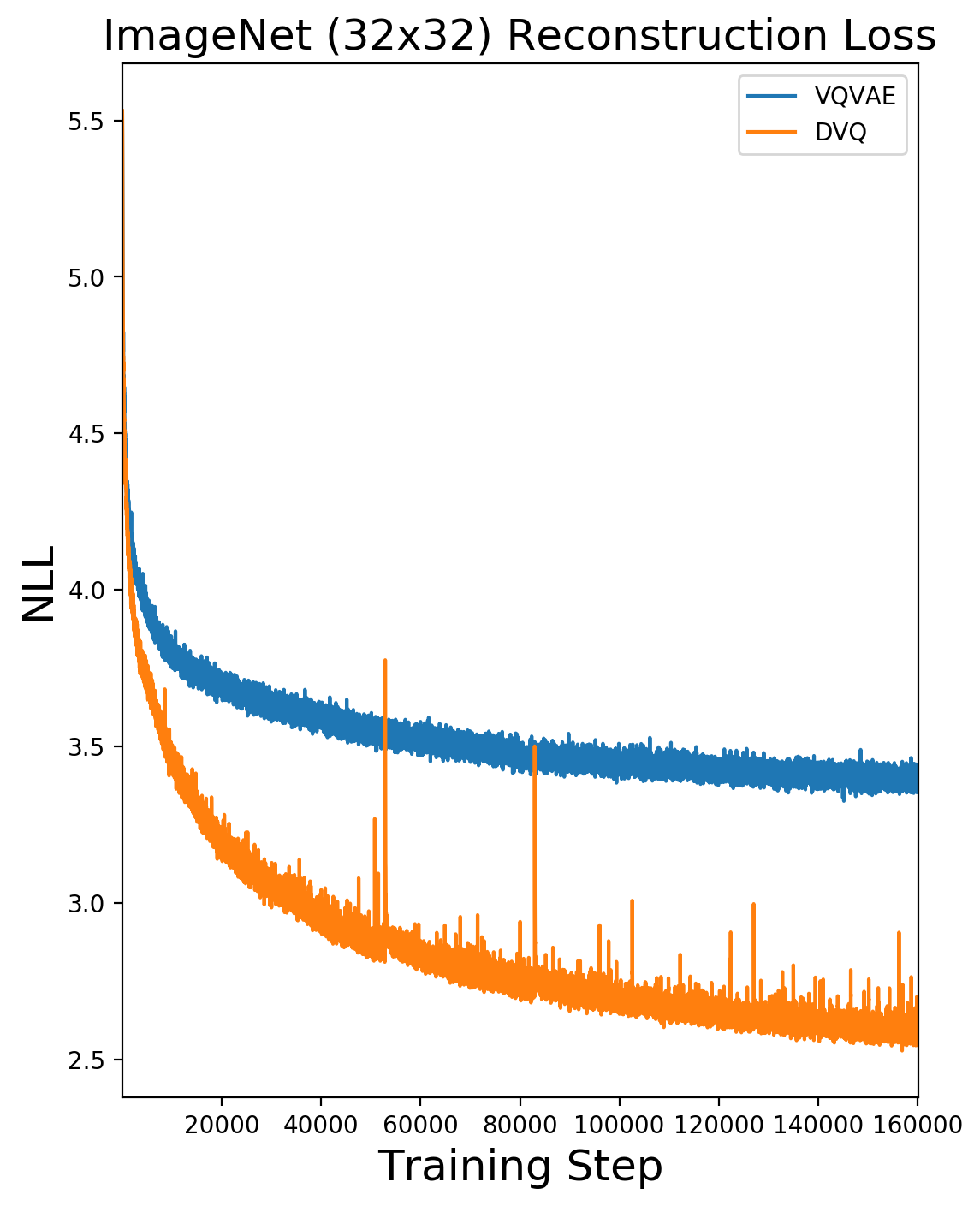}
\end{subfigure}
\begin{subfigure}{.3\textwidth}
\includegraphics[width=\textwidth]{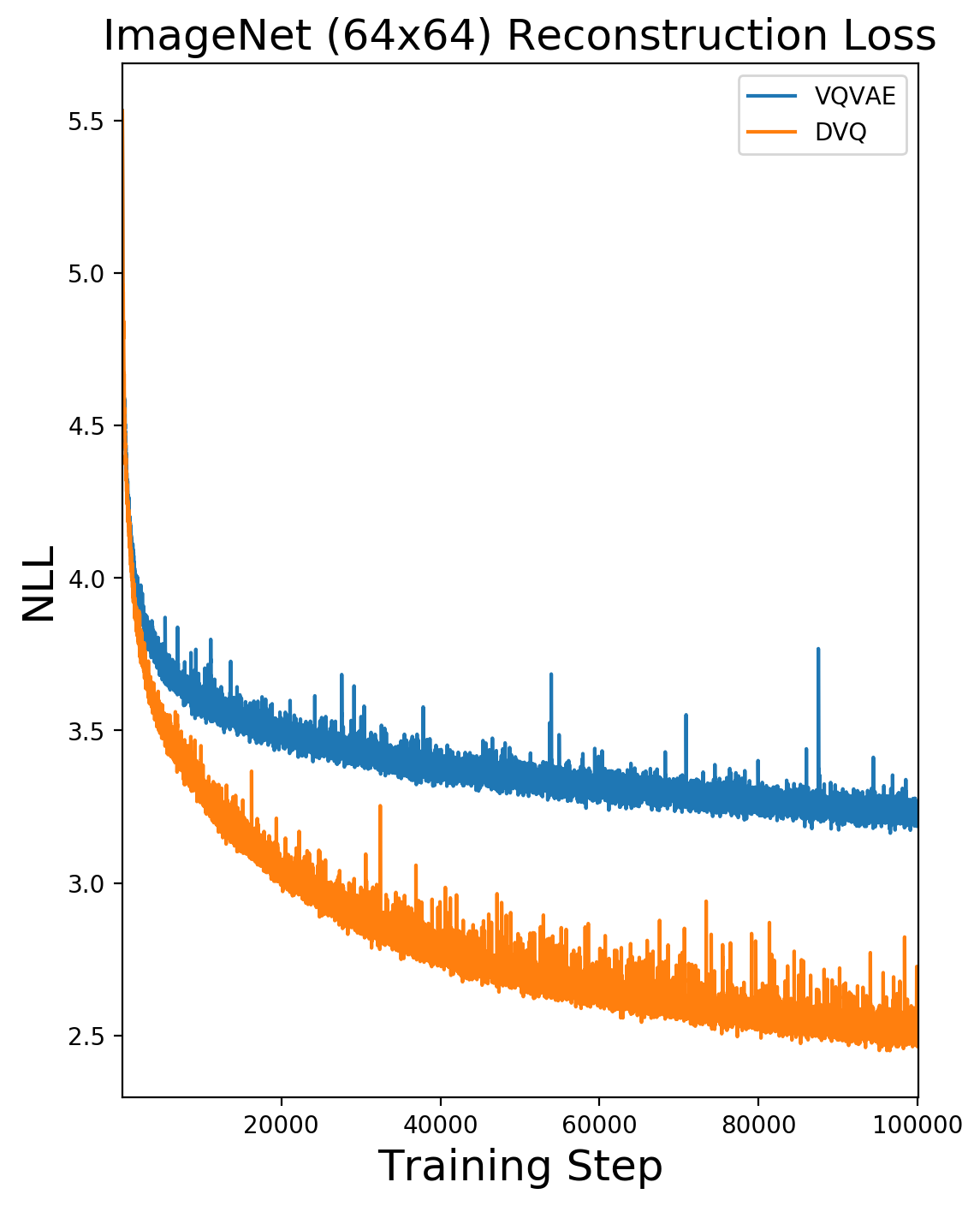}
\end{subfigure}
\begin{subfigure}{.3\textwidth}
\includegraphics[width=\textwidth]{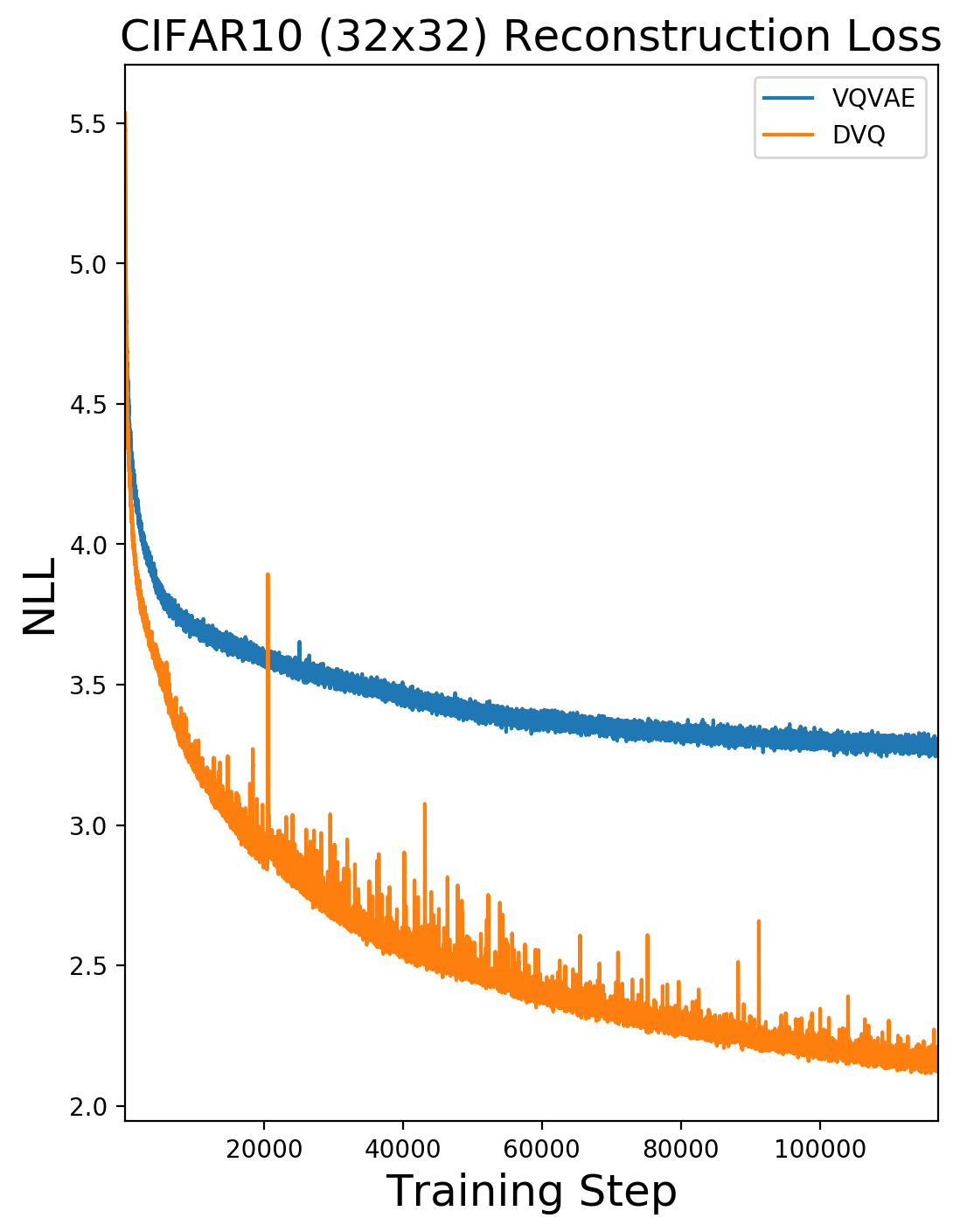}
\end{subfigure}
\caption{Reconstruction loss while training on different resolutions of ImageNet and CIFAR10 using $L=10$}\label{loss_imagenet_figures}
\end{figure}

The advantage is not only on the final reconstruction loss. As it is shown on Figure \ref{loss_imagenet_figures}, DVQ converges at a much faster rate then VQVAE.

In table \ref{benchmarks} you can see the benchmarks for VQVAE for ImageNet 32x32 and 64x64, compared to DVQ and VQVAE. DVQ performs significantly better than VQVAE and at par with auto-regressive models.

\begin{figure}
\begin{subfigure}{0.5\textwidth}

\includegraphics[width=\textwidth]{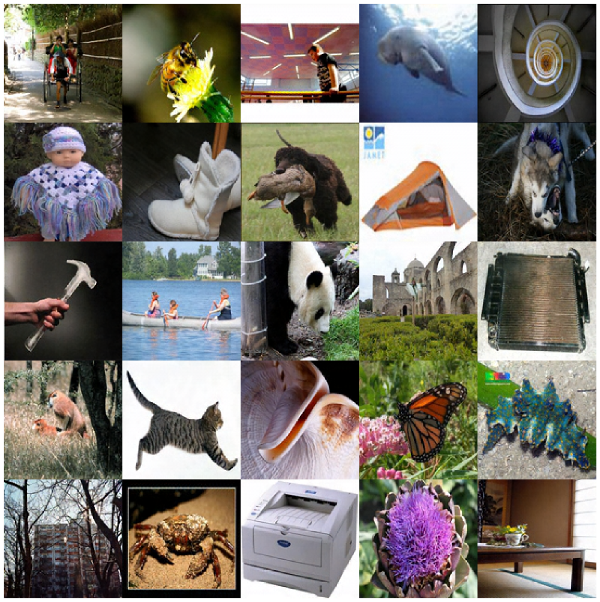}
\end{subfigure}
\begin{subfigure}{0.5\textwidth}
\includegraphics[width=\textwidth]{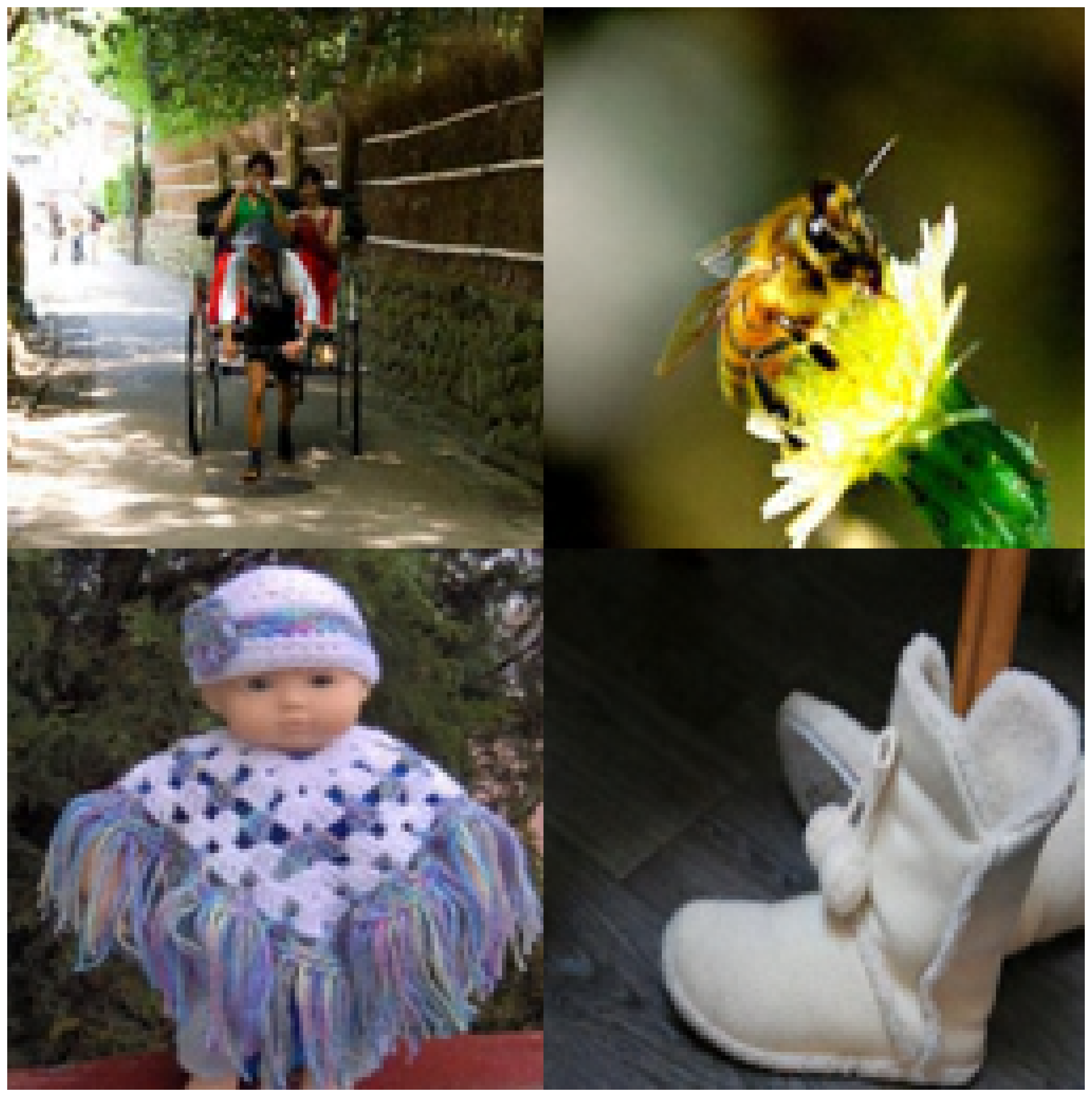}
\label{fig:imgnet128}
\end{subfigure}

\caption{Reconstructions on ImageNet 128x128 with $L=10$}
\end{figure}

\begin{table}
\caption{Reported benchmark in bits/dim for different models. We report all our benchmarks for the mean loss in the test data.}\label{benchmarks}
\begin{tabularx}{\textwidth}{|X|X|X|X|}
\hline
Model &  CIFAR-10 & ImageNet(32x32) & ImageNet(64x64)\\
\hline
 VQVAE  \cite{van2017neural} &  4.67  &4.92 & 4.66 \\
 DVQ (Ours) & \textbf{3.15}  & \textbf{3.76} & \textbf{3.50}\\
SVQ (Baseline) & 5.85  &- & -\\
Gated PixelCNN &3.03 & 3.83 &3.57 \\
\hline
\end{tabularx}
\end{table}

\section{Related Work}
There is no work to the best of our knowledge that proposes depthwise or feature wise application of multiple vector quantizers in a context of deep architectures or density estimation with VQ. In this section we discuss previous work that is most relevant to our work in three important aspects. 
\begin{itemize}
 \item    Works that consider the implications of applying transformations depthwise (accross feature space) as opposed to spatially or temporally  
    
  \item   Works that consider improvements in vector quantizers and mitigation of problems in performance
  \item  Works that introduce improvements in discrete representation learning
 \end{itemize}

\subsection{Ensemble Models}
There are several ensemble models techniques that consider features independently when training. One example is that of Random Forests. Such techniques train multiple models on a subset of features as opposed to multiple models jointly. Ensemble models can often get improved training performance and surpass that of their individually trained components. 

\subsection{Xception}
Xception\cite{chollet2017xception} based architectures can obtain improved results as compared to the original work (Inception\cite{szegedy2015going}) by applying convolutions depthwise, across feature space as opposed to spatially. 

\subsection{Vector Quantization}
Recent improvements in VQ focus on computational efficiency. LSQ++ \cite{martinez2018lsq++} propose parallelism advantages on training in disjoint subsets of a dataset. However they don't consider the performance advantages of training on subsets of feature space. 

Other early works \cite{juang1982multiple} in Vector Quantization have proposed multiple vector quantizers applied together. Such architectures, when considered alone or in combination with other components such as HMM \cite{segura1994multiple}, they are applied on the same joint feature space. 

We couldn't find any works, recent or past that have presented and evaluated results by directly considering codebooks on different subsets of the feature space.

Other works propose applying multiple codebooks on different slices along the time axis of a timeseries (such as different time steps of audio) \cite{burton1983generalization}.

\subsection{VQVAE}

Derivative works on VQVAE introduce improvements by applying VQ in a hierarchical fashion at multiple levels of the encoder and decoder network \cite{razavi2019generating}. Such works are able to learn discrete representations on different granularity of feature space at each level and produce remarkably high resolution images. However such works do not consider applying the VQ on the feature axis at a given level. Roy et al \cite{roy2018theory} trains a VQ using an EM objective. Their approach leads to improved training and inference performance.

\section{Conclusions}
Our work can be applied in multiple architectures that use Vector Quantization, including but not limited to generative works that train auto-regressive models on the discretized prior. One challenge in training an autoregressive model in the prior distribution is that that there are multiple codebooks with independent depthwise features. Some auto-regressive models such as PixelCNN++ reject the independence assumption in favor of performance. Thus choosing the correct experimental set up might be a challenge.  We would like to see our work applied in that context and evaluated. Moreover we would like to see the parallelism performance benefits by dividing the feature space in a similar fashion as LSQ++ \cite{martinez2018lsq++}


We conclude that our work provides an advantage because our assumptions on feature independence is partially correct in the context we evalauted. Such as images and features extracted by CNN. We would like to see when this assumption is wrong, that is when features are highly dependent, and how this would affect the performance of our approach. Our code is available at: \url{https://github.com/Paper-Under-Review/DVQ}

\bigskip \noindent\textbf{Acknowledgements.} We would like to thank Dr. Greg Ver Steeg, Sami Abu-El-Haija and Chin-Cheng Hsu for the helpful discussions.
%
%
%
\bibliographystyle{splncs04}
\bibliography{bib}
%




\newpage
\section{Appendix}

\subsection{Additional Reconstructions}
\begin{figure}[H]
\begin{subfigure}{.5\textwidth}
\includegraphics[width=\textwidth]{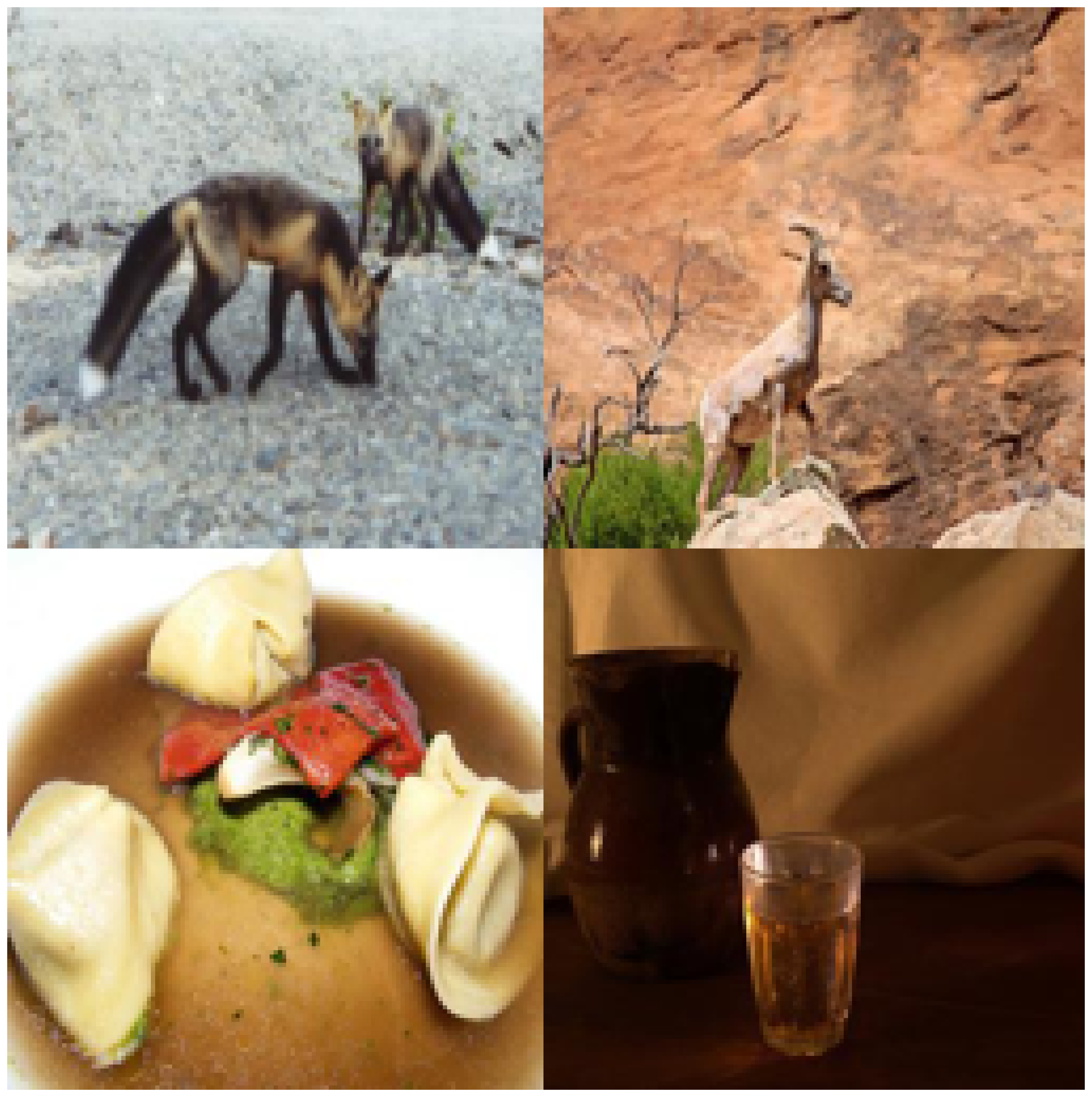}
\end{subfigure}
\begin{subfigure}{.5\textwidth}
\includegraphics[width=\textwidth]{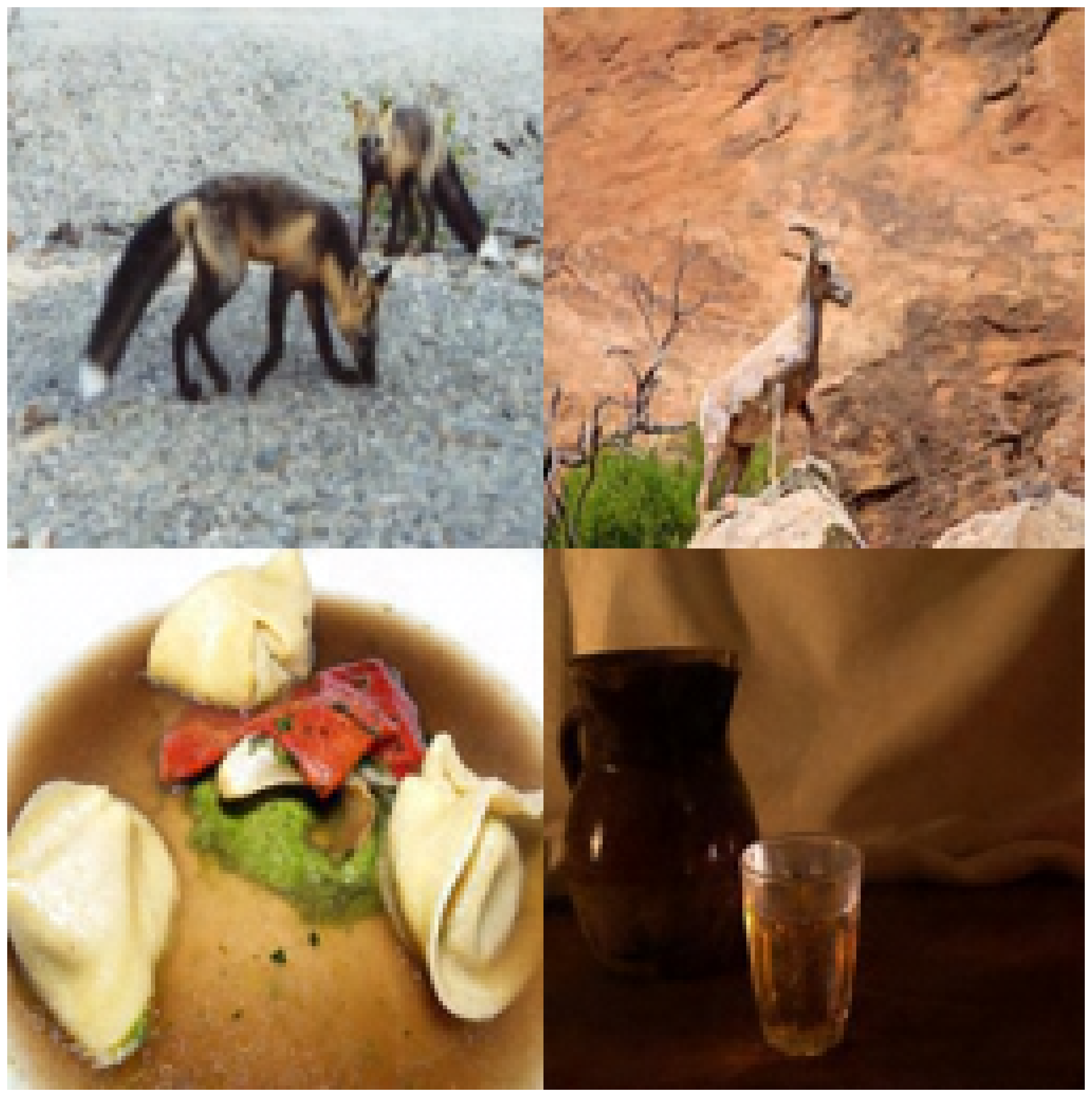}
\end{subfigure}
\begin{subfigure}{.5\textwidth}
\includegraphics[width=\textwidth]{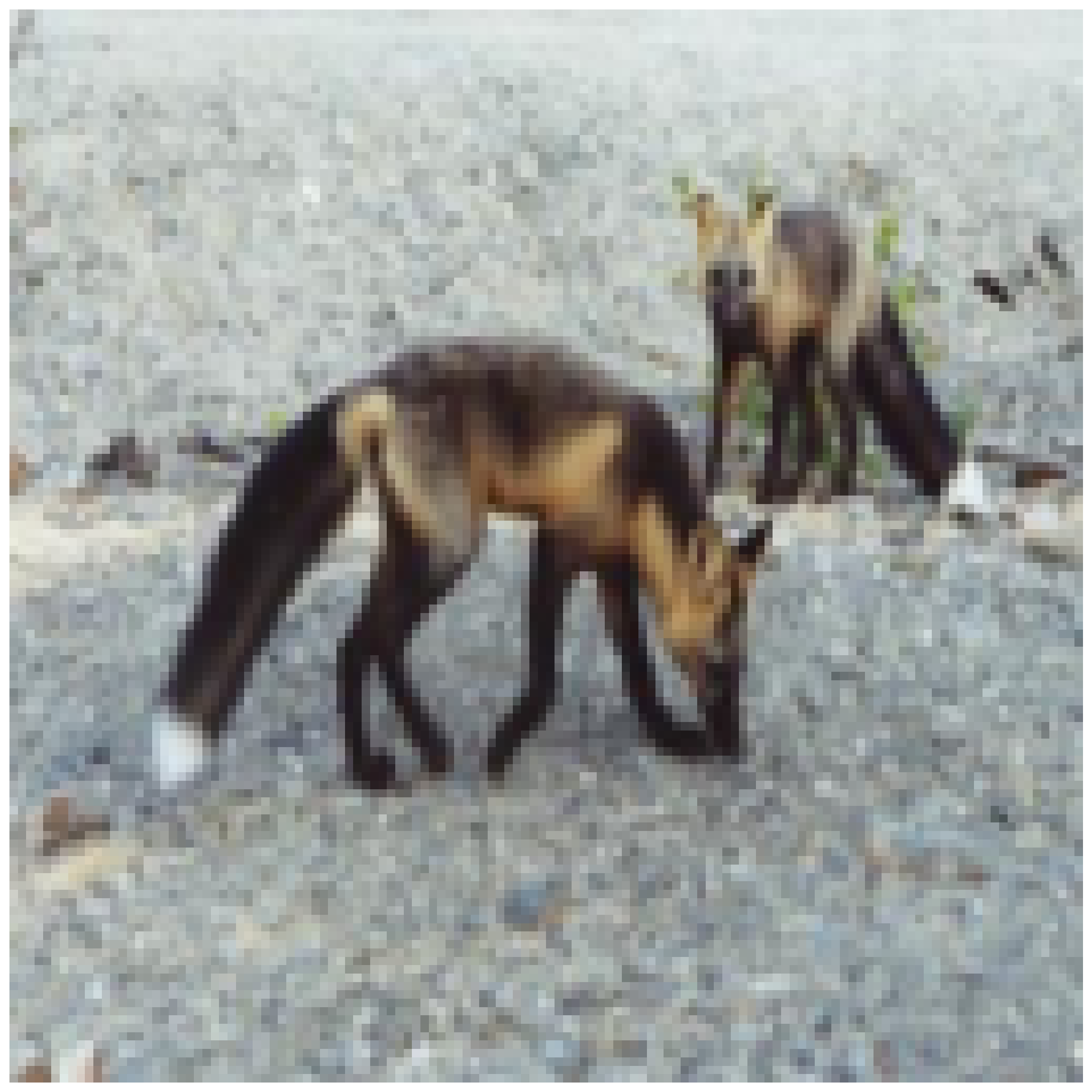}
\end{subfigure}
\begin{subfigure}{.5\textwidth}
\includegraphics[width=\textwidth]{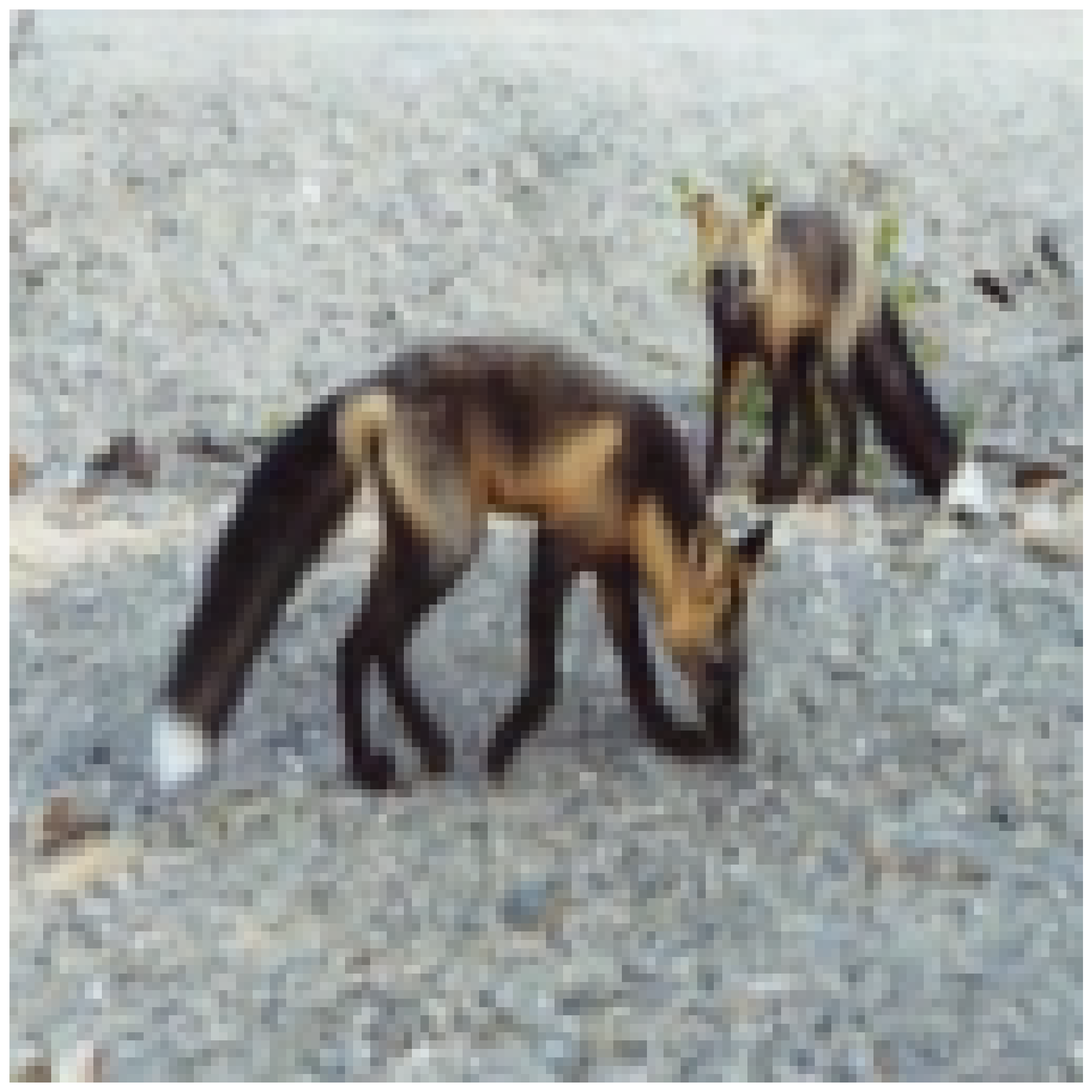}
\end{subfigure}
\caption{Left: Original Image Right: Reconstruction. ImageNet Test set 128x128}
\end{figure}
\begin{figure}[H]
\includegraphics[width=\textwidth]{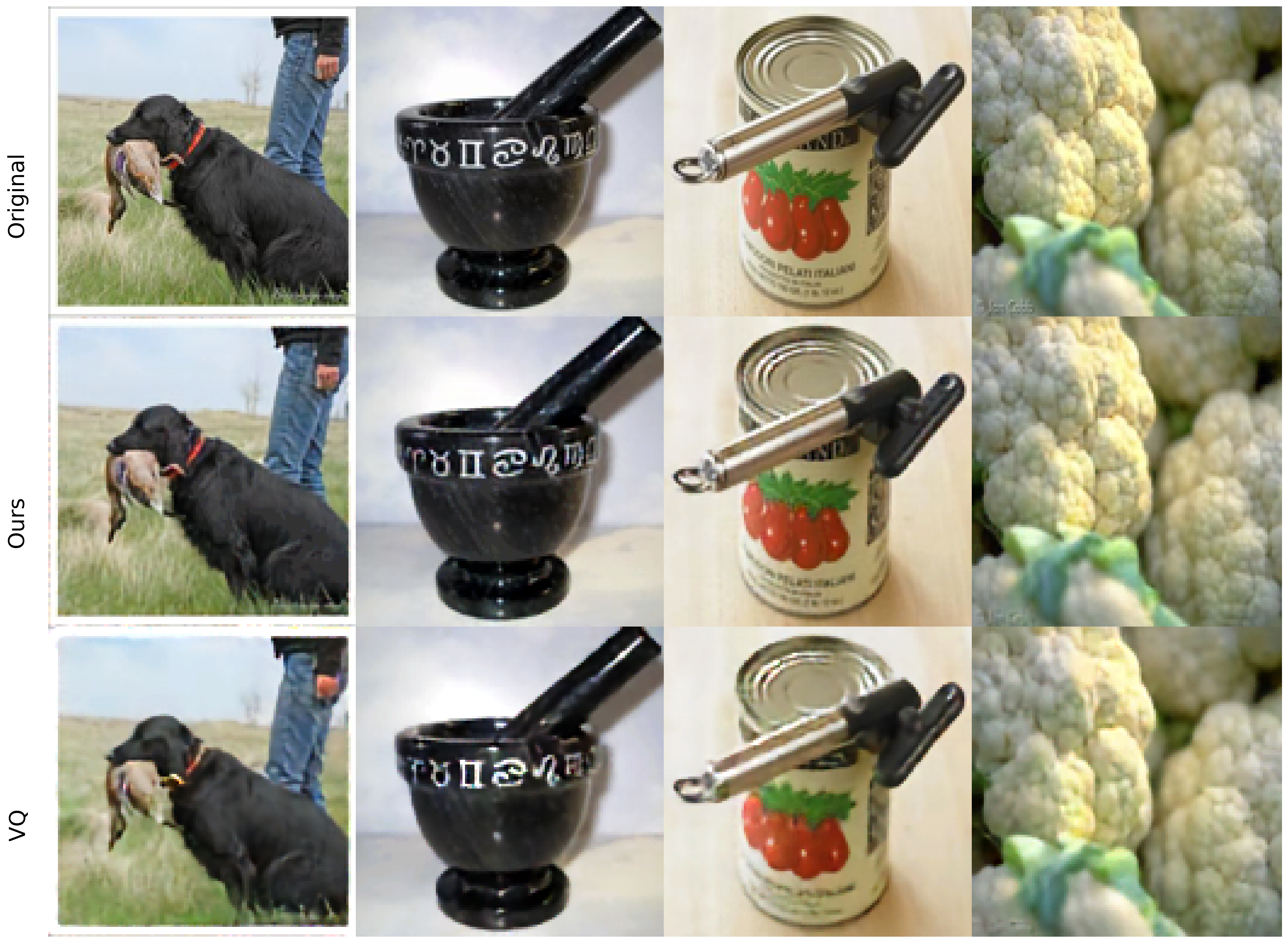}
\caption{ImageNet 128x128 Reconstructions (NLL Model) Comparison with VQ}
\end{figure}
\begin{figure}[H]
\includegraphics[width=\textwidth]{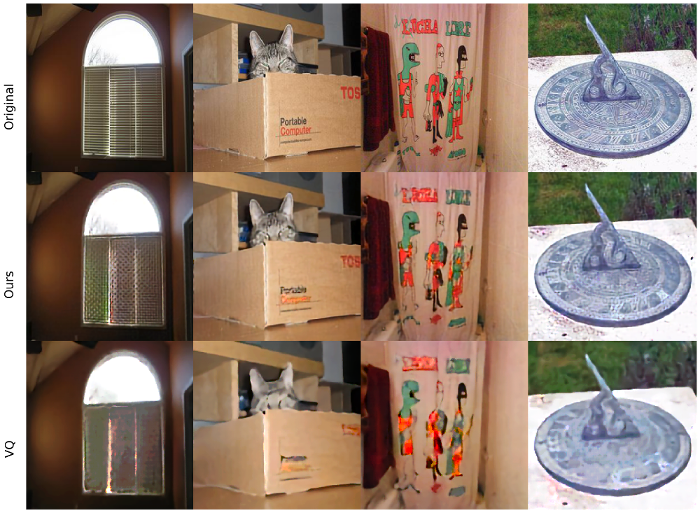}
\caption{ImageNet 256x256 Reconstructions (NLL Model) Comparison with VQ}
\end{figure}
\subsection{Network Architecture}

\begin{figure}[H]
\includegraphics[width=\textwidth]{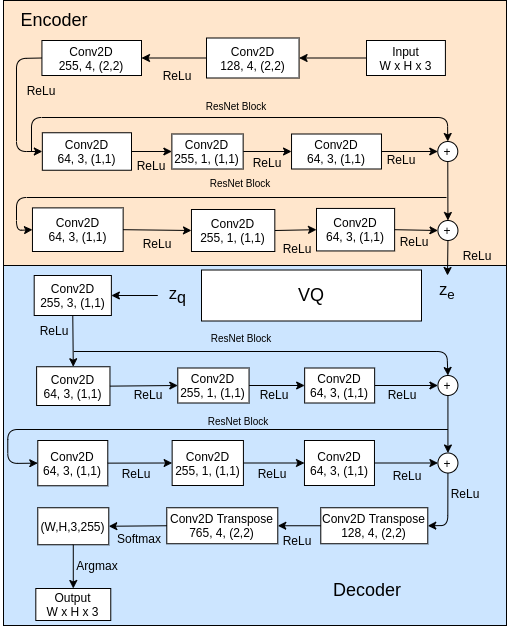}
\caption{Network Architectural Overview}
\end{figure}

\subsection{Static Prior Analysis}

\begin{figure}[H]
\includegraphics[width=\textwidth]{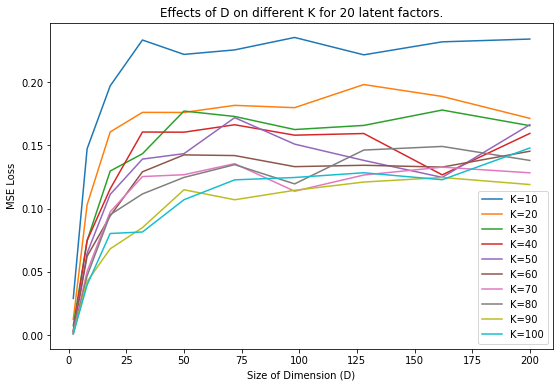}
\caption{K becomes irrelevant when dimensionality of feature space is too large}
\end{figure}
\begin{figure}[H]
\includegraphics[width=\textwidth]{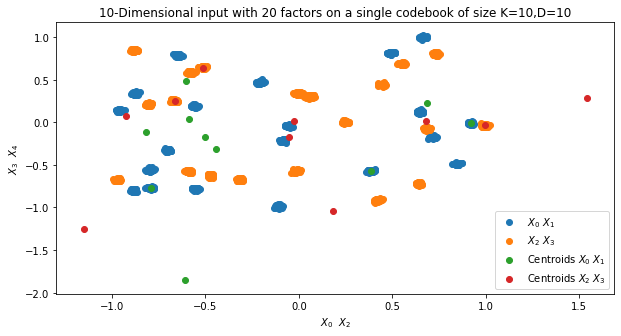}
\caption{Visual Inspection of Density estimation by Multiple Codebooks. Red centroids correspond to Orange Densities and Green to Blue. When K is too small, some centroids attempt to cover multiple clusters. }
\end{figure}
\begin{figure}[H]
\includegraphics[width=\textwidth]{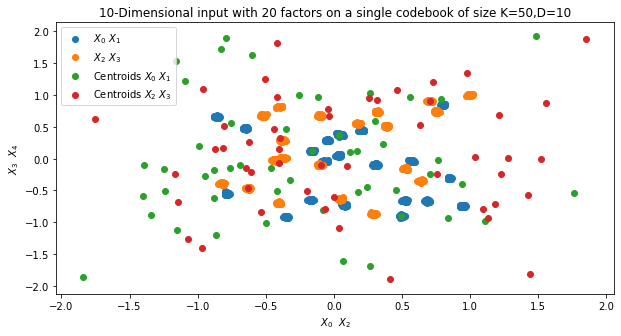}
\caption{Visual Inspection of Density estimation by Multiple Codebooks. Red centroids correspond to Orange Densities and Green to Blue. When K is too large it doesn't always lead to an optimal mapping for high dimensional spaces}
\end{figure}
\end{document}